\def\year{2019}\relax
\documentclass[letterpaper]{article} 
\usepackage{aaai19}  
\usepackage{times}  
\usepackage{helvet}  
\usepackage{courier}  
\usepackage{url}  
\usepackage{graphicx}  

\usepackage{subcaption} 
\usepackage{multirow}
\usepackage[table]{xcolor}
\usepackage[hang,flushmargin]{footmisc}

\nocopyright

\begin{document}
%
\title{Quantifying Human Behavior on the Block Design Test
Through Automated Multi-Level Analysis of Overhead Video}
\author{Seunghwan Cha\thanks{Present affiliation: Hong Kong University of Science and Tech.}, James Ainooson, and Maithilee Kunda \\
        Electrical Engineering and Computer Science, Vanderbilt University, Nashville TN, USA \\
        scha@connect.ust.hk, james.ainooson@vanderbilt.edu, mkunda@vanderbilt.edu
}
\maketitle
\begin{abstract}
The block design test is a standardized, widely used neuropsychological assessment of visuospatial reasoning that involves a person recreating a series of given designs out of a set of colored blocks.  In current testing procedures, an expert neuropsychologist observes a person's accuracy and completion time as well as overall impressions of the person's problem-solving procedures, errors, etc., thus obtaining a holistic though subjective and often qualitative view of the person's cognitive processes.  We propose a new framework that combines room sensors and AI techniques to augment the information available to neuropsychologists from block design and similar tabletop assessments. In particular, a ceiling-mounted camera captures an overhead view of the table surface. From this video, we demonstrate how automated classification using machine learning can produce a frame-level description of the state of the block task and the person's actions over the course of each test problem.  We also show how a sequence-comparison algorithm can classify one individual's problem-solving strategy relative to a database of simulated strategies, and how these quantitative results can be visualized for use by neuropsychologists.
\end{abstract}

\section{Introduction}

\begin{figure}[h]
    \centering
    \includegraphics[width=0.6 \linewidth]{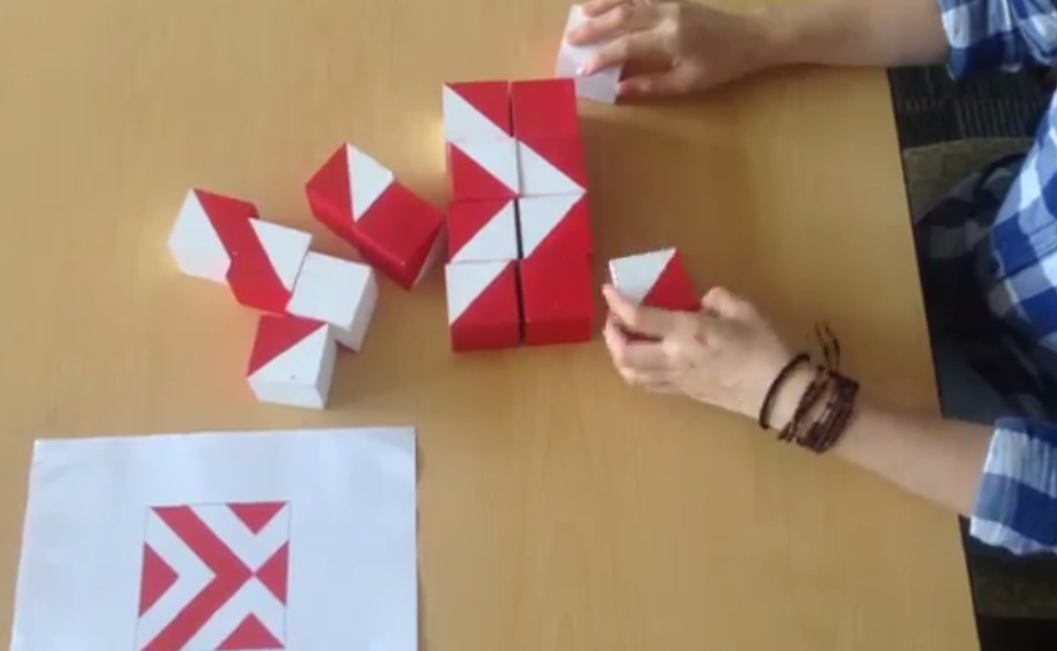}
    \caption{A person solving a sample block design item.  (To protect test security, actual test items are not shown.)} \label{fig:BD_photo}
    \vspace{-6pt}
\end{figure}

The block design test (BDT) is a widely-used neuropsychological assessment of visuospatial reasoning that forms a part of most standardized IQ tests \cite{wechsler2003wechsler} and is often used to characterize cognition in individuals with developmental or neurological conditions \cite{lezak2004neuropsychological}.  In the BDT, a person has to reconstruct a given printed design using red and white blocks, as shown in Figure \ref{fig:BD_photo}.  

\begin{figure}[t]
    \centering
    \includegraphics[width=\linewidth]{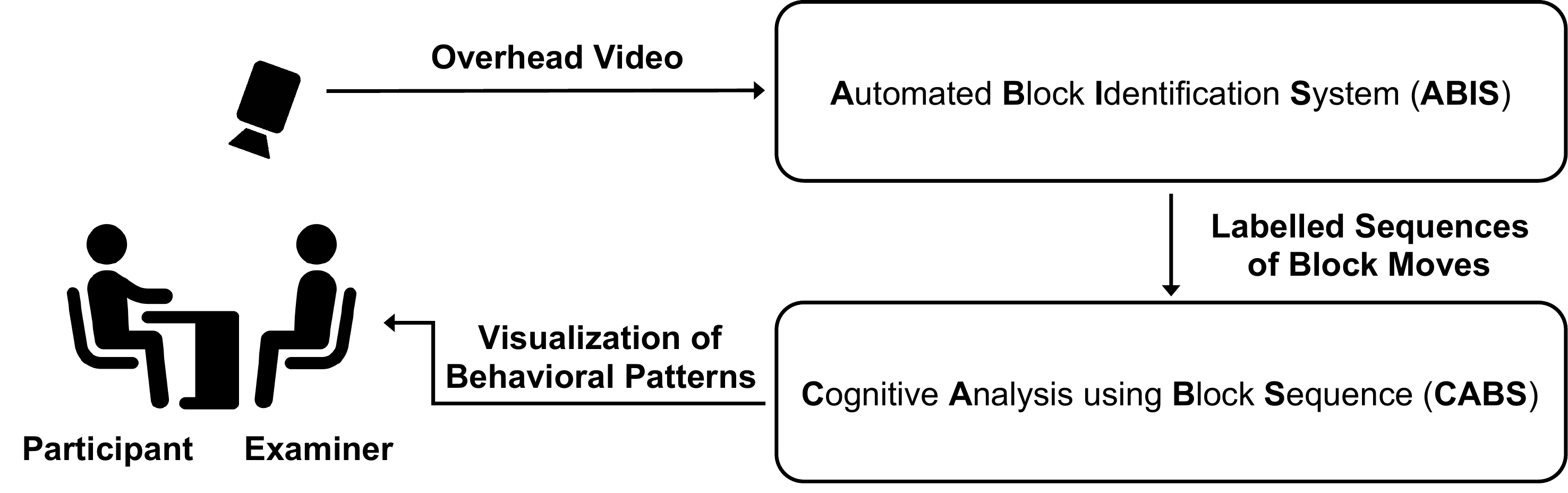}
    \caption{Framework for using automated analyses of overhead video to augment information about human block design performance available to neuropsychologists. 
    }
    \label{fig:summary}
\end{figure}

\begin{figure*}[t]
    \centering
    \includegraphics[width=0.98\linewidth]{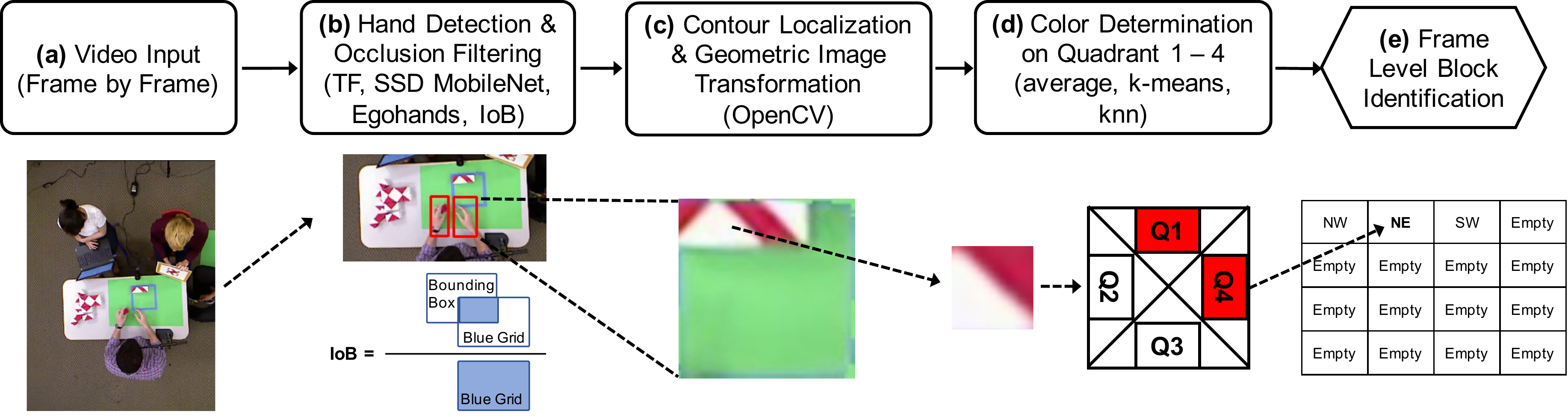}
    \caption{Details of Automated Block Identification System (ABIS). \textbf{(a)} Frames of the overhead video recording are given as the input. \textbf{(b)} Hand occlusions are filtered out through hand detection using the Tensorflow Object Detection API \cite{Huang_2017_CVPR}. The SSD Mobilenet network \cite{Liu2017_ssd} was pre-trained on COCO and re-trained for hand detection on the Egohands Dataset \cite{Bambach_2015_ICCV}. Frames having Intersection over Blue (IoB) greater than 0.3 were filtered out. \textbf{(c)} The blue contour was localized and transformed through OpenCV geometric image transformation functions \cite{opencv_library}. \textbf{(d)} The upright image from the previous step is then divided into n x n block-sized sub-images.  Sub-images are again divided into 4 quadrants. The colors of each quadrant are used to obtain the final block label, as shown in \textbf{(e)}.}
    \label{fig:abis_summary}
\end{figure*}

The BDT is typically scored using the test-taker's final accuracy and reaction time.  However, many other aspects of a person's BDT performance can be highly informative about their cognitive characteristics.
For example, the original BDT scoring system included tallying the \textit{number of block moves made} \cite{kohs1920block}, but this was later deemed too cumbersome for practice \cite{hutt1932kohs}.
BDT errors have been studied in terms of \textit{particular sequences of block moves} \cite{joy2001quantifying,rozencwajg2002strategy,toraldo2004error}, \textit{incorrect placements of blocks} \cite{ben1971similarities,hoffman2003spatial,jones1981analysis,joy2001quantifying,schatz2000hierarchical,troyer1994age}, and \textit{qualitative types or scale of errors} \cite{akshoomoff1989block,akshoomoff1996influence,joy2001quantifying,kramer1991visual,kramer1999configural,schatz2000hierarchical,troyer1994age,zipf2000qualitative}.

\textbf{Despite the known value of having these kinds of rich behavioral measurements from the BDT, such information is rarely collected in practice,} mainly due to the difficulty a neuropsychologist would have in recording such data accurately and in real time, while also administering the test, gathering basic scores, and attending to other important aspects of the testing session \cite{milberg2009boston}.

To solve this problem, we propose a new, AI-based framework for automatically measuring behavior in the context of the BDT and similar tabletop cognitive assessments. Figure \ref{fig:summary} illustrates the two systems that comprise our framework: (1) the Automated Block Identification System (ABIS) uses overhead video to extract a frame-by-frame description of the state of the block task, capturing all block moves, errors, etc made by the test-taker; and (2) the Cognitive Analysis using Block Sequence (CABS) system uses information about specific block sequences to summarize and visualize pertinent information for the neuropsychologist.

Here, we present a proof-of-concept evaluation of this framework using video data recorded from a human participant study conducted with college students.  In future work, we aim to expand the capabilities of these systems and also conduct human studies with larger and more varied samples.

\section{Participant Study and Data Collection}

Our team conducted a participant study to collect data about human performance on the Block Design Test (BDT).  All necessary IRB approvals were obtained for this research.

The full study included 14 adult participants from a college student population.  Participants completed the full BDT from the Wechsler IQ scales \cite{wechsler2003wechsler} plus four additional sample puzzles that we created.  The room contained an overhead Kinect RGB-D camera, and participant gaze was also measured using various eye-tracking technologies.  The setup included covering the table with a green sheet and using larger-than-standard blocks (see Figure \ref{fig:abis_summary}a) in order to facilitate initial research progress with our AI systems.  Note that on the BDT, all blocks are identical and have three pairs of distinct faces (all red, all white, and diagonal red/white).

In this paper, we analyze a subset of data from the full study: we focus on data from the overhead camera, RGB only; we consider seven participants (labeled here as A-G) for whom these videos were fully annotated at the time of this writing; and we focus on six BDT puzzles (two from the original BDT, which we cannot depict here as they are protected testing materials, plus the four sample puzzles that we created and can show alongside our analyses).

Overhead videos were annotated using the ELAN software tool \cite{wittenburg2006elan}.  For each participant completing each BDT puzzle, annotations included individual block placement locations (position within the blue square outline that we created to guide participants' block construction activities) and block faces (Empty, Red, White, Northwest (NW), Northeast (NE), Southwest (SW), Southeast (SE)), where directional labels indicate a diagonal red/white block face with the red half pointing in the specified direction).  Annotations also included transition periods during which any block was in motion.  For this initial study, annotations were completed by a single research team member.

We had to omit data from one participant (participant B, puzzle 6) because this participant showed a highly irregular block placement strategy requiring a modified annotation scheme.  Thus, for this study, we use data from 41 trials (six puzzles across seven participants, minus one omitted trial).

\section{Part 1: Automated Block Identification System}

The Automated Block Identification System (ABIS) takes an overhead video recording as input, and outputs a description of the state of the tabletop at every frame.  Figure \ref{fig:abis_summary} gives an overview of the operation of the ABIS system. 

\textbf{Locating block positions.} The first challenge in detecting the state of the tabletop is to accurately locate the block positions from the overhead video.  Though the overhead camera was fixed in our study, movements of the physical table and of the green sheets on the table meant that the position of the blue ``construction area'' could change from trial to trial.  

As shown in Figure \ref{fig:abis_summary}, the blue contour was located using blue HSV ranges, from the start of each trial before any blocks were placed. The green area within the blue contour was also detected using color ranges to crop out the blue tape outline. The coordinates of the cropped image in the initial frame were saved and used throughout the rest of the trial.  Since the blue grid was not always perfectly aligned with the image frame, the cropped image was rotated to an upright position using standard geometric image transformations in the OpenCV library \cite{opencv_library}. 

Finally, this cropped and rotated image of the ``construction area'' was divided into n x n sub-images, depending on the size of the give BDT puzzle.  In the sample BDT puzzles studied here, two puzzles were of size 3 x 3 (total of 9 block spaces) and four puzzles were of size 4 x 4 (total of 16 block spaces).  Each of these block-sized sub-images was then fed into a color determination function.

\textbf{Block color determination.}
The block-sized sub-images were further divided into 4 quadrants. By looking at various if-then combinations of the measured color in each quadrant, we can classify a block face into categories of Empty, Red, White, Northwest (NW), Northeast (NE), Southwest (SW), and Southeast (SE), as described in the data collection section.  We also add a label of ``Invalid'' if none of the labels seem to fit the observed colors.
For instance, as shown in Figure \ref{fig:abis_summary}d, if quadrants 1 and 4 are red while quadrants 2 and 3 are white, the sub-image is classified as Northeast (NE). 

We experimented with several techniques to recognize the color of individual block-face quadrants:

\textit{RGB Averaging.}  In our first method, each quadrant's pixel values were averaged in each respective RGB channel and compared with a threshold value of 140. This threshold was set empirically based on initial experiments. 

\textit{K-Means Clustering.}  We also used clustering on the pixel values of the three color channels to acquire the dominant RGB value in the quadrant \cite{Kanungo00anefficient}. We tested this approach with k-values of 1 and 4, and kept the same threshold value of 140.

\textit{K-Nearest Neighbors.}  The final color recognition method we tried was K-Nearest Neighbors (KNN) \cite{knn}. Red, white, and green color images (10 each) served as the training data for this color recognition task. Color histograms of each color channel and the corresponding color were recorded to train a classifier. With the trained classifier, the quadrant images and color histograms of those images were used as test data. For this approach, instead of comparing a computed value of pixels with the threshold level, the KNN classifier returned the color of the quadrant directly.

\textbf{Hand detection and filtering.}
Based on initial results from these color determination methods (as detailed in the following section), we hypothesized that occlusions by participants' hands might be affecting color determination performance, so we experimented with methods to detect hands and filter out frames in which hands occlude the block construction area.

As an initial approach, we simply used available annotations about when any block was in motion, and excluded all such frames from block identification results. 

We also used standard deep learning-based methods to detect hands and determine whether the hand bounding box did overlap the construction area.  Among the many available options, we chose to use the Single Shot Multibox Detector (SSD) MobileNet network available from the Tensorflow Object Detection API \cite{Huang_2017_CVPR}, pre-trained on the COCO dataset \cite{Liu2017_ssd} and then re-trained for hand detection using the EgoHands dataset \cite{Bambach_2015_ICCV}.

Hand detection results were (qualitatively) not excellent, likely because the Egohands dataset used for retraining only contains egocentric hand views, whereas our videos contain overhead views.  To improve detection performance, we had to lower the detection threshold, increase the number of total hands detected in an image, and restrict the search perimeter to the table where all the relevant hand movement occurred.  After these adjustments, we often got redundant detections on a single hand, but these errors did not affect our final results, as we only needed rough estimates of hand positions.

Finally, given hand positions defined by bounding boxes, we calculated what we called the \textit{Intersection over Blue grid (IoB)} value, i.e., how much do hands overlap the block construction area, as shown in Figure \ref{fig:abis_summary}b.  If the IoB was larger than 0.3, we set the current block state to be equal to the previous frame's block state. By copying the result from the previous frame, the occlusion effect on the detection can essentially be filtered out. The threshold of 0.3 was set to preserve frames where the hand bounding box overlaps with the construction area only slightly.

\textbf{Across-frame smoothing.}
Finally, we applied a smoothing technique to reduce discontinuities in the calculated block labels. In particular, as a participant's hand creates an occlusion, some locations where blocks have already been identified become newly classified as \textit{Invalid}. After the hand moves away from that location, the classification result return to the original label. Such temporary misclassifications can be removed by smoothing out any \textit{Invalid} labels that occur between two identical labels for a given block position. 

\subsection{Part 1 Results}

\begin{table}[!b]
\fontsize{5pt}{8pt}\selectfont
\caption{Accuracy results for different settings.}\label{tab1}
\resizebox{\columnwidth}{!}{
    \begin{tabular}{c|r|r}
    \hline
    Category & Parameter Setting & Accuracy \\
    \hline
    \multirow{4}{*}{Color Determination}
                                            & RGB Averaging & 0.68 \\
                                            & K-Means Clustering (k=1) & 0.68 \\
                                        & K-Means Clustering (k=4) & 0.64 \\
                                        & K-Nearest Neighbors & 0.67 \\
    \hline
    \multirow{2}{*}{Hand Filtering}   & Block in Motion & 0.73\\
                                        & Hand Bounding Box & 0.69 \\
    \hline
    Post Processing                     & Smoothing & \textbf{0.81}\\
    \hline
\end{tabular}
}
\label{table:accuracy_table}
\end{table}

Overall results from our ABIS experiments are shown in Table \ref{table:accuracy_table}.  We used the manual ELAN annotations of block identities as the ground truth.

The best method for color determination ended up being simple RGB Averaging.  We expect this is because of small image size; the overall image frames were only 480 x 640, and as shown in Figure \ref{fig:abis_summary}a, the blue grid itself was very small within this image.  As a result, each quadrant used to perform color determination was only about 3 x 3 pixels in size. In the future, we plan to test other methods such as neural network approaches for the color determination step, and also collect data using higher resolution cameras.

\begin{table}[htbp]
\fontsize{5pt}{8pt}\selectfont
\caption{Accuracy results across all participants and puzzles.
}\label{tab2}
\resizebox{\columnwidth}{!}{\begin{tabular}{c|c|c|c|c|c|c|c|c}
\hline
Participant & A & B & C & D & E & F & G & Avg\\
\hline
Puzzle 1 & \cellcolor{green!12}0.78 & \cellcolor{green!68}0.92 & \cellcolor{green!72}0.93 & \cellcolor{green!12}0.78 & \cellcolor{green!24}0.81 & \cellcolor{red!8}0.73 & \cellcolor{green!60}0.90 & 0.84 \\
Puzzle 2 & \cellcolor{red!24}0.69 & \cellcolor{red!60}0.60 & \cellcolor{green!56}0.89 & \cellcolor{green!64}0.91 & \cellcolor{red!12}0.72 & \cellcolor{green!52}0.88 & \cellcolor{green!60}0.90 & 0.80 \\
Puzzle 3 & \cellcolor{green!68}0.92 & \cellcolor{green!4}0.76 & \cellcolor{green!52}0.88 & \cellcolor{green!64}0.91 & \cellcolor{green!68}0.92 & \cellcolor{green!64}0.91 & \cellcolor{green!28}0.82 & 0.87 \\
Puzzle 4 & \cellcolor{green!24}0.81 & \cellcolor{red!24}0.69 & \cellcolor{green!36}0.84 & \cellcolor{green!96}0.99 & \cellcolor{green!40}0.85 & \cellcolor{green!84}0.96 & \cellcolor{red!8}0.73 & 0.84 \\
Puzzle 5 & \cellcolor{red!56}0.61 & \cellcolor{green!32}0.83 & \cellcolor{green!52}0.89 & \cellcolor{red!8}0.73 & \cellcolor{red!80}0.55 & \cellcolor{red!28}0.68 & \cellcolor{red!20}0.70 & 0.71 \\
Puzzle 6 & \cellcolor{green!12}0.78 & \cellcolor{black!18}n/a & \cellcolor{green!8}0.77 & \cellcolor{red!12}0.72 & \cellcolor{red!4}0.74 & \cellcolor{green!60}0.90 & \cellcolor{green!44}0.86 & 0.80  \\
\hline
Avg       & 0.77 & 0.76 & 0.87 & 0.84 & 0.77 & 0.84 & 0.82 & \textbf{0.81} \\
\hline
\end{tabular}}
\label{table:detail_table}
\end{table}


Accuracy improved slightly through hand occlusion filtering, as shown in the second section of Table \ref{tab1}, which gives results using the RGB Averaging color determination method.  Removal of manually annotated frames labeled as having blocks in motion resulted in better accuracy than the automated hand detection method, but we found that quite a large number of frames had been removed.  After removal, only around 20\% of the original frames remained, because participants are almost constantly moving blocks around.  Further work on this hand detection step is needed.  

\begin{figure}[t]
	\centering
	\includegraphics[width=0.95\linewidth]{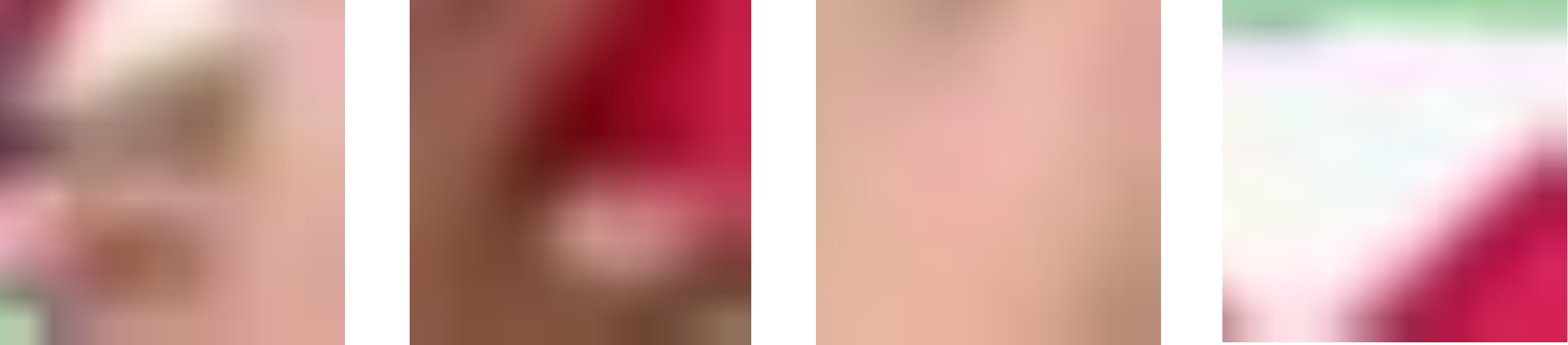}
		\caption{Examples of misclassification by our block identification system.  From left to right, calculated labels were: \textit{SW, Red, White, White}.}
		\label{fig:failure}
\end{figure}

Finally, the smoothing step significantly improved the average accuracy by more than 10\%.  The smoothing result in Table \ref{tab1} uses RGB Averaging and the automated hand detection methods described above.

Table \ref{table:detail_table} illustrates ABIS accuracy for each participant and puzzle, using RGB Averaging, automated hand detection, and smoothing.  There are interesting variations across participants and puzzles, though we are still working to find the root causes of these variations.  Some participants may have been more meticulous in their block placements than others, which would be expected to yield higher accuracies by the ABIS system.  In addition, some individual trials may have been affected by shifts in the table or green table covering, which would negatively affect accuracy.  

Figure \ref{fig:failure} depicts some failure cases of our ABIS classifier.  In addition to hand-occlusion-related misclassifications, we also found errors when the geometric image transformation step failed to accurately identify or align the block construction area prior to dividing it into block-sized sub-images.

\section{Part 2: Cognitive Analysis of Block Sequences}

Given a block move sequence, how can a person's Block Design Test (BDT) performance be further quantified?  The Cognitive Analysis using Block Sequence (CABS) system is a tool for summarizing and visualizing patterns in individual BDT strategies.  Here, we present examples of CABS functionalities for demonstration purposes.  Ultimately, such functionalities would be designed to capture behavioral features of relevance identified by the psychological literature and through discussions with neuropsychologists.

\begin{figure}[b]
    \centering
    \includegraphics[width=\linewidth]{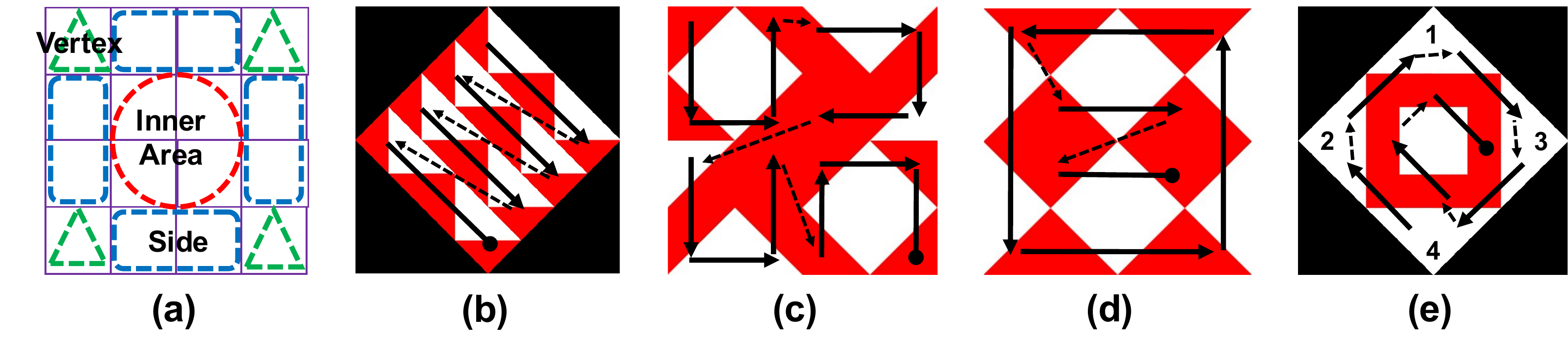}
    \caption{\textbf{(a)} Regions of the block construction area, labeled as vertices, sides, and the inner area. \textbf{(b-e)} Examples of spatial block placement strategies, with arrows representing the order of blocks placed: \textbf{(b)} row-by-row, \textbf{(c)} sub-section, \textbf{(d)} perimeter-complete, and \textbf{(e)} vertices-first.
    }
    \label{fig:strategy-terminology}
\end{figure}

\begin{figure*}[t]
    \centering
    \includegraphics[width=0.98\linewidth]{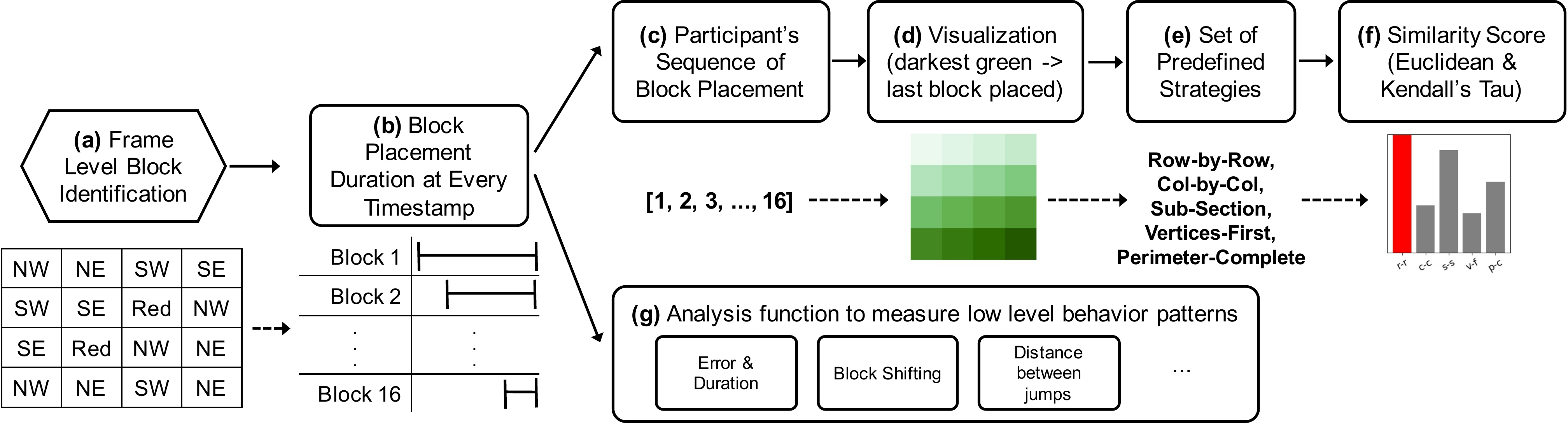}
    \caption{Details of Cognitive Analysis using Block Sequence (CABS). \textbf{(a)} Frame level block identification is given as an input from ABIS. \textbf{(b)} Block placement duration can be obtained at every timestamp which represents the duration of particular block label at certain block location. \textbf{(c-d)} Participant's sequence of block placement can be used for visualization to better understand one's block solving strategy. This sequence represents the order in which each block location was first tackled. Later modification to the already placed block is not accounted for. \textbf{(e-f)} Database of simulated strategies are prepared and compared with the participant's sequence through use of either Euclidean distance or Kendall's Tau. \textbf{(g)} Another functionality is to measure low level behavior patterns of a participant. Examples of such features include correlation between error and duration and block shifting.}
    \label{fig:cabs_summary}
\end{figure*}

\begin{figure*}[!t]
    \centering
    \includegraphics[width=\linewidth]{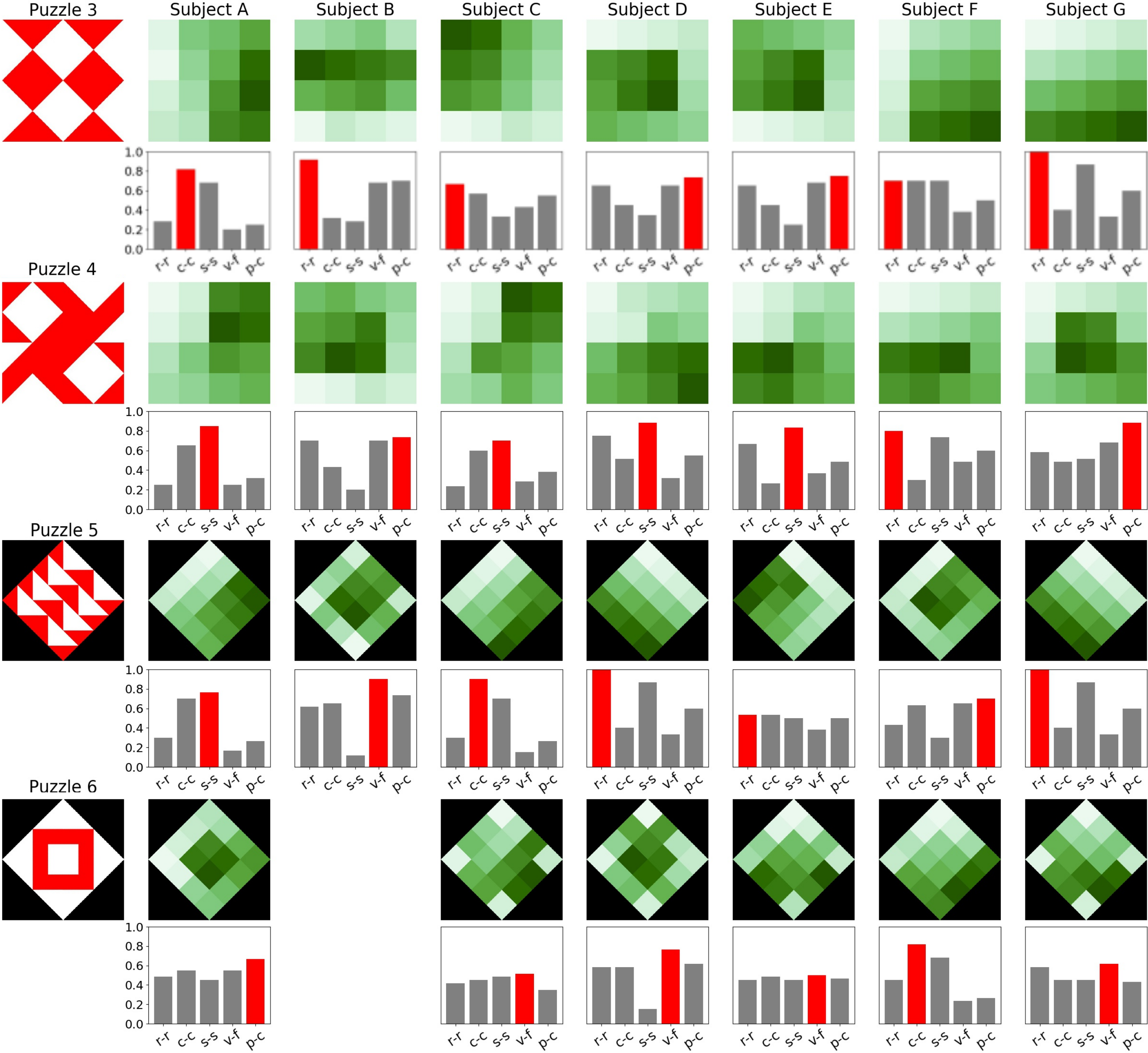}
    \caption{Visualization of each subject's block solving strategy from puzzle 3 to 6. Only 4 x 4 puzzles were presented here as we need to include the puzzle design. The sequence is encoded in the color gradient where lightest green means first move while darkest green shows last move. Subject's strategy can be determined by comparing Kendall's tau coefficient with the true lists of predefined strategies: \textit{row-by-row (r-r), column-by-column (c-c), sub-section (s-s), vertices-first (v-f), and perimeter-complete (p-c)}. Most likely strategy which is the one with highest similarity score is colored red in the graph below.}
    \label{fig:strategy-visualization}
\end{figure*}

Figure \ref{fig:cabs_summary} shows an overview of the CABS system.  First, frame-level block identifications are used to derive a block placement sequence, i.e., when each individual block was placed in each position in the block construction area.  Repeated placements within the same position usually indicate errors that the test-taker made and later fixed.

Then, we use this block sequence information to derive two types of analysis: 1) classification of an individual's block placement sequence relative to a predefined set of spatial strategies; and 2) extraction of low level behavior patterns.  For this initial study, we use sequence data obtained directly from the manual annotations of participant recordings. In the future, as our ABIS detection algorithms mature, we will feed ABIS outputs directly into the CABS system.

\subsection{Strategy classification}

From initial inspection of our participant study results, we observed that there seemed to be some overall trends in the high-level ``spatial block placement strategies'' used by various participants.  For example, some participants would complete puzzles systematically, going from one block row to the next.  Other participants demonstrated a more disordered approach, placing blocks at various disjoint locations.  

The first CABS functionality that we implemented aimed to classify individual block placement sequences according to their similarity across a collection of predefined strategies.  (Note that we define sequences here according to initial block placements; later modifications of blocks are not incorporated into these sequence analyses.  Block placement errors are addressed in the following subsection, on CABS calculations of other low-level behavioral features) 

We first defined five block placement strategies that were (somewhat) observed in our participant data: row-by-row (r-r); column-by-column (c-c); sub-section (s-s); perimeter-complete (p-c); and vertices-first (v-f).  Figure \ref{fig:strategy-terminology}a illustrates the regions of the 4 x 4 block construction area that these strategies involve, and Figures \ref{fig:strategy-terminology}b-e illustrate four of these strategies (column-by-column is not shown, as it is similar to row-by-row).  Note that the CABS approach presented here would work for any arbitrary collection of strategies; we use these five just as examples to demonstrate the approach.

Then, using this predefined collection of strategies as a starting point, we developed a method to classify an individual's block placement sequence on a given puzzle according to its similarity across this collection of strategies. Participant sequences rarely matched a ``pure'' version of any of these strategies, and the possible number of individual sequence variations are combinatorially very large.  Thus, instead of calculating similarity analytically, we calculate similarity empirically, by creating a ``sample set'' of concrete sequence examples for each abstract strategy type, and then compute similarity of an individual's actual sequence across each sample set of sequences.  There are even too many variations of ``pure'' sequences of each strategy type to fully enumerate (in some cases numbering in the millions), and so we created 576 (4! x 4!) strategies in each sample set for the five different strategy types.  We created these sample sets by manually writing scripts that would generate valid sequences by iterating across different possible starting positions, etc.

Then, similarity scores were calculated by comparing the sequence information from the participant to all sequences in each sample set. We used two different similarity calculations.  Similarity was first computed by calculating the Euclidean distance between two lists and normalizing this distance using \cite{Qian:2004:SEC:967900.968151}:
$$euc = 1 / (1 + dist(x, y)) $$ 
\noindent In this equation, $x$ and $y$ represent corresponding elements from the two lists and \textit{dist} is the standard Euclidean distance function.  After experimenting with this similarity formulation, we realized that this approach has disadvantages, mainly because sequence differences become weighted by the magnitude of numbering differences, which are essentially arbitrary in the context of comparing two block placement sequences. 

So instead, we adopted a second approach to use the Kendall's tau coefficient, which evaluates the degree of similarity between two sets of ranked lists \cite{kendall1938measure}. Kendall's tau can be calculated as:
$$tau = (P - Q) / sqrt((P + Q + T) * (P + Q + U))$$
\noindent where P is matching pairs, Q is non-matching pairs, T is the number of ties only in x, and U is the number of ties only in y. Values close to 1 indicate strong agreement between two ranked lists whereas values near -1 indicate strong disagreement, i.e., -1 implies complete rank inversion.  The Kendall's tau method yielded results that were overall similar to the Euclidean distance based method, but without the undesirable edge cases that would be possible with the latter.

Finally, using the Kendall's tau equation given above, similarity scores were calculated for an individual block placement sequence relative to all sequences in each sample set.  The resulting maximum similarity across the sample set is output as the final similarity score for a participant's sequence and each of the five predefined strategies (as shown in Figure \ref{fig:strategy-terminology}).

Results from this block sequence analysis are illustrated in Figure \ref{fig:strategy-visualization}.  On the left side of each row, the target design is shown for four different sample BDT puzzles, and the seven remaining images to the right of each row are visualizations of individual participant's block placement sequences, encoded using a color gradient to represent time. The lightest green represents the earliest block that was placed, and the darkest green represents the last block. 

Even by itself, this visualization can help observers quickly grasp similarities and differences in the block placement strategies used by different participants. For example, participants C and G followed similar patterns for solving puzzle 6, while participant A showed a very different strategy for the same puzzle.

The bar graphs below each sequence visualization show similarity scores for that individual sequence across the five predefined strategies: row-by-row (r-r); column-by-column (c-c); sub-section (s-s); perimeter-complete (p-c); and vertices-first (v-f) (as shown in Figure \ref{fig:strategy-terminology}).  
Graphs with all bars at roughly the same height indicate mixed or unspecified strategies, i.e., the sequence did not show particularly noteworthy similarity to one strategy type over another.  Graphs in which one bar dominates the others indicates a stronger match to a particular strategy.  A bar with value of 1.0 indicates a perfect match between the participant's sequence and at least one of the sequences in the sample sets.

It is interesting to note that each participant does not adhere to a single block placement strategy throughout different puzzle. Instead, they seem inclined to find the best strategy for each puzzle. Some common trends were that the vertices-first strategy only appeared in puzzles 5 and 6, which are both diagonally oriented puzzles, and the sub-section was used the most for puzzle 4, which has strong symmetric characteristics.  Thus, the choice of strategy is likely closely related to the shape of the target design, which matches with findings from the BDT literature \cite{caron2006cognitive,farran2001block,kramer1999configural,miller2009learning,stewart2009autistic} and poses many interesting questions for future work.

\subsection{Low level behavioral features}

In addition to analyzing block placement sequences and strategies, the CABS also contains functions to extract and visualize other low-level behavioral features that would be of interest to neuropsychologists.  There are many interesting features that can be obtained just from the block sequence information input to CABS, including, for example:
\begin{itemize}
    \item Number and types of errors.
    \item Intermediate and final reaction times.
    \item Spatial distance between consecutive block placements.
    \item Progression tendencies, e.g., left to right vs. right-to-left.
    \item Single versus multiple simultaneous block placements.
    \item Block pair swapping versus in-place block changes.
    \item Correlations among any/all of the above.
\end{itemize}

While a fully functioning CABS would require input from neuropsychologists and the BDT literature to define an appropriate set of features, we present here an initial look at some of the individual variations that occurred in our participant data, even though our study was conducted with a fairly homogeneous sample (adult college students, mostly from a couple of different majors).


In particular, our CABS system can output an error analysis for each participant by detecting changes in the block label at a given position from one time to another.  For instance, if a participant placed a \textit{NW} block in a certain position but later on changed the block to \textit{NE}, CABS would be able to detect that change as an error.

\begin{figure}[b]
    \centering
    \includegraphics[width=\linewidth]{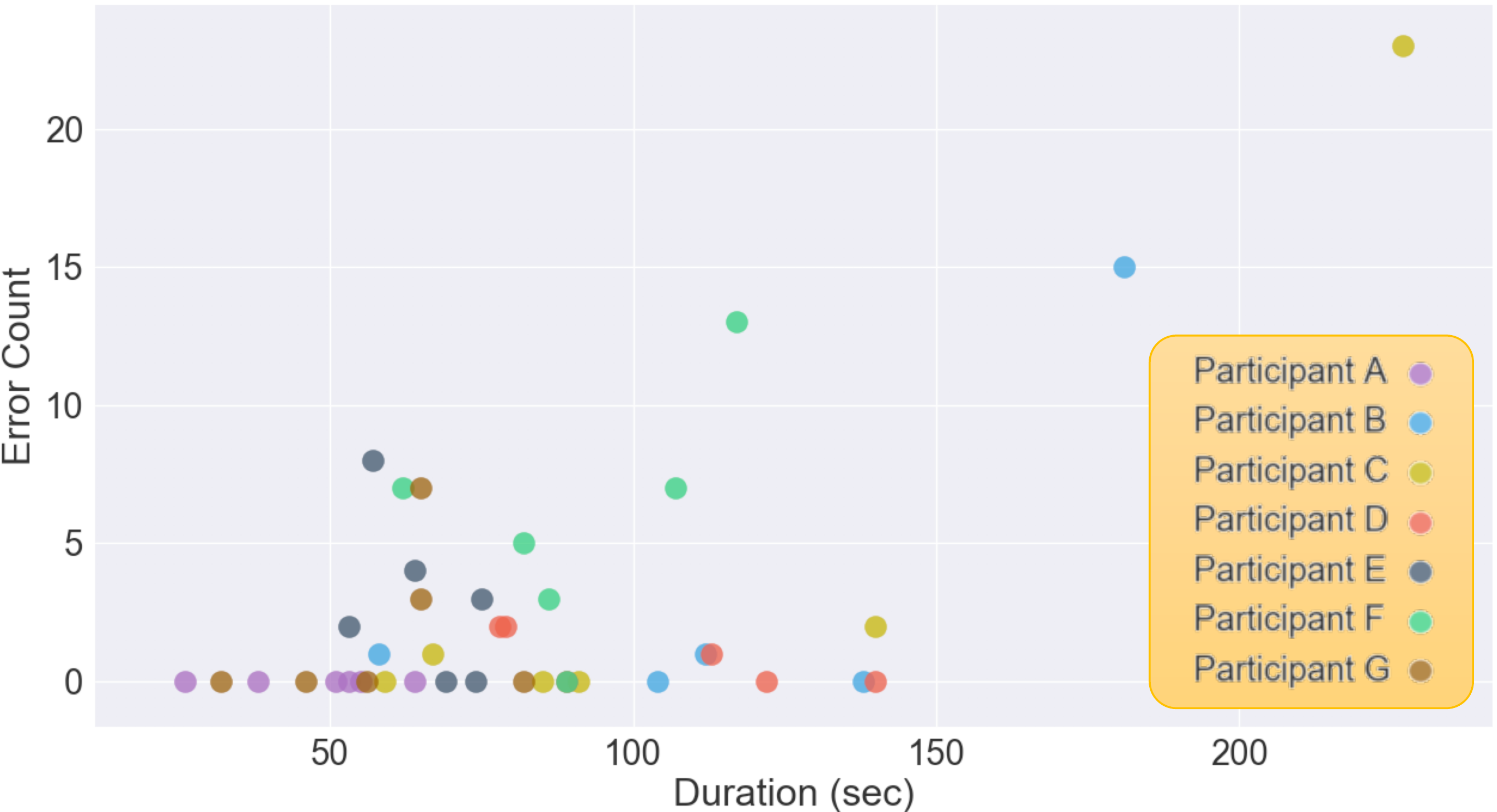}
    \caption{Error count and duration correlation graph.}
    \label{fig:error-correlation}
\end{figure}

As illustrated in Figure \ref{fig:error-correlation}, we can generate a scatter plot to visualize relationships between CABS error counts and overall puzzle completion times across participants in a sample group.  Each participant's results are labeled using different colors, with one marker dot per individual puzzle completed. There are several interesting patterns in this plot.  

For instance, participant A generally finished much earlier than participant B and also made far fewer errors throughout. Participant C did not make a lot of errors in general, but struggled greatly on one puzzle, making more than 20 errors and taking more than 200 seconds to complete the design, a clear outlier across other participants and puzzles.

Such visualizations could be of use both to researchers studying the BDT as well as practicing neuropsychologists.  Test practitioners could easily compare an individual to normative groups, or compare individual performance across multiple time points, for example to help in detecting cognitive decline in the elderly.

\section{Conclusion and Future Work}

In this paper, we presented a proof-of-concept demonstration of a new, AI-based framework to aid in the automated analysis of human behavior on the Block Design Task or other widely-used tabletop neuropsychological assessments.  This framework contained the Automated Block Identification System (ABIS) that used computer vision and machine learning techniques to accurately detect block placements at the per-frame level.  A second system, the Cognitive Analysis using Block Sequence (CABS) system, contained various functionalities for extracting and visualizing information about participants' problem solving strategies for use by neuropsychologists or other test practitioners.

\begin{figure}[t]
    \centering
    \includegraphics[width=\linewidth]{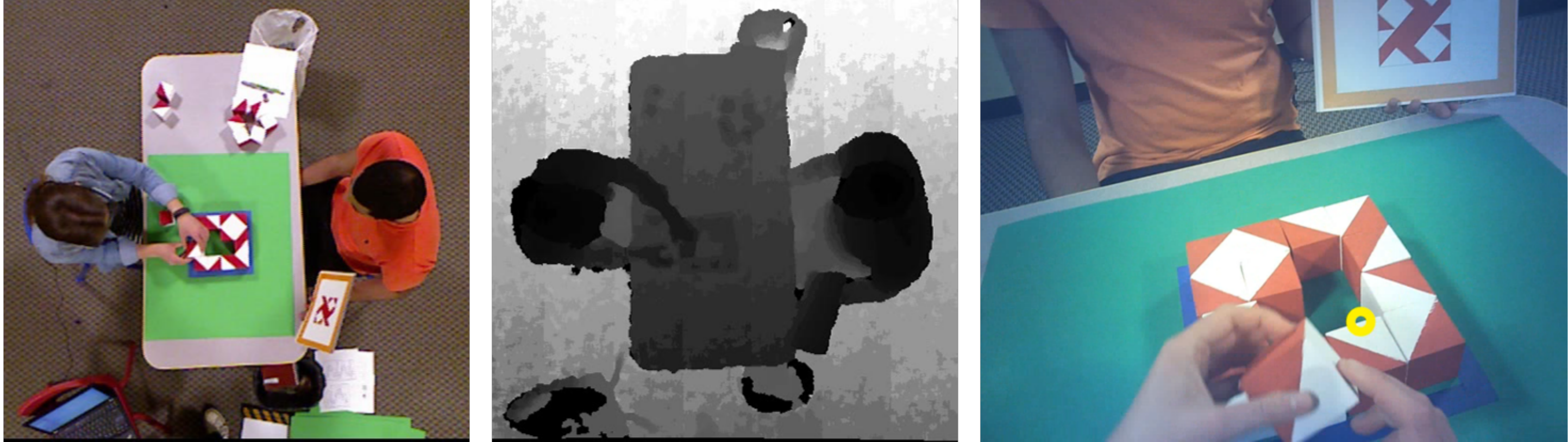}
    \caption{A top down RGB image (left). The same snapshot showing depth information (middle). Snapshot from a wearable eye-tracker mounted on the participant's head, with a yellow circle indicating the gaze target in the scene (right).}
    \label{fig:depth}
\end{figure}

Our results, while promising, do highlight many areas for improvement. For example, as illustrated in Part 1, more robust detection algorithms must be developed for the ABIS in order to overcome vision problems related to the misalignment of block images, hand occlusions, etc.  Such problems would become more severe in real block design test administrations, which use smaller blocks, have no fixed grid on the table, and no green background.  In Part 2, continued development of the CABS requires ongoing interactions with neuropsychologists, in order to ensure that the system can capture and measure behavioral features that are of clinical relevance.

In addition, while this paper focused on analyses using just an overhead RGB video, the incorporation of additional sensor modalities will also greatly expand the quality and types of analyses of human BDT performance that can be obtained. As illustrated in Figure \ref{fig:depth} (middle), the use of depth information could be useful for more accurate automated hand and block detection.  Head-mounted wearable eye trackers can provide valuable information about a participant's cognitive processes as they solve the test.

However, adding more sensors such as these has little practical value unless the ensuing data can be analyzed automatically and reliably, and the results of such analyses can be presented in an usable format to the intended users.  Continued research on systems like the ABIS and CABS presented here will help to translate algorithmic and technological advances into much-needed practical applications in neuropsychology and related clinical settings.

\clearpage
\bibliography{references}

\begin{thebibliography}{}

\bibitem[\protect\citeauthoryear{Akshoomoff and
  Stiles}{1996}]{akshoomoff1996influence}
Akshoomoff, N.~A., and Stiles, J.
\newblock 1996.
\newblock The influence of pattern type on children's block design performance.
\newblock {\em J. International Neuropsychological Society} 2(5):392--402.

\bibitem[\protect\citeauthoryear{Akshoomoff, Delis, and
  Kiefner}{1989}]{akshoomoff1989block}
Akshoomoff, N.~A.; Delis, D.~C.; and Kiefner, M.~G.
\newblock 1989.
\newblock Block constructions of chronic alcoholic and unilateral brain-damaged
  patients: A test of the right hemisphere vulnerability hypothesis of
  alcoholism.
\newblock {\em Archives of Clinical Neuropsychology} 4(3):275--281.

\bibitem[\protect\citeauthoryear{Bambach \bgroup et al\mbox.\egroup
  }{2015}]{Bambach_2015_ICCV}
Bambach, S.; Lee, S.; Crandall, D.~J.; and Yu, C.
\newblock 2015.
\newblock Lending a hand: Detecting hands and recognizing activities in complex
  egocentric interactions.
\newblock In {\em The IEEE International Conference on Computer Vision (ICCV)}.

\bibitem[\protect\citeauthoryear{Ben-Yishay \bgroup et al\mbox.\egroup
  }{1971}]{ben1971similarities}
Ben-Yishay, Y.; Diller, L.; Mandleberg, I.; Gordon, W.; and Gerstman, L.
\newblock 1971.
\newblock Similarities and differences in block design performance between
  older normal and brain-injured persons: A task analysis.
\newblock {\em J. Abnormal Psychology} 78:17--25.

\bibitem[\protect\citeauthoryear{Bradski}{2000}]{opencv_library}
Bradski, G.
\newblock 2000.
\newblock {The OpenCV Library}.
\newblock {\em Dr. Dobb's Journal of Software Tools}.

\bibitem[\protect\citeauthoryear{Caron \bgroup et al\mbox.\egroup
  }{2006}]{caron2006cognitive}
Caron, M.; Mottron, L.; Berthiaume, C.; and Dawson, M.
\newblock 2006.
\newblock Cognitive mechanisms, specificity and neural underpinnings of
  visuospatial peaks in autism.
\newblock {\em Brain} 129(7):1789--1802.

\bibitem[\protect\citeauthoryear{Cover and Hart}{1967}]{knn}
Cover, T., and Hart, P.
\newblock 1967.
\newblock Nearest neighbor pattern classification.
\newblock {\em IEEE Transactions on Information Theory} 13(1):21--27.

\bibitem[\protect\citeauthoryear{Farran, Jarrold, and
  Gathercole}{2001}]{farran2001block}
Farran, E.~K.; Jarrold, C.; and Gathercole, S.~E.
\newblock 2001.
\newblock Block design performance in the {W}illiams syndrome phenotype: A
  problem with mental imagery?
\newblock {\em J. Child Psychology and Psychiatry} 42(6):719--728.

\bibitem[\protect\citeauthoryear{Hoffman, Landau, and
  Pagani}{2003}]{hoffman2003spatial}
Hoffman, J.~E.; Landau, B.; and Pagani, B.
\newblock 2003.
\newblock Spatial breakdown in spatial construction: Evidence from eye
  fixations in children with {W}illiams syndrome.
\newblock {\em Cognitive Psychology} 46(3):260--301.

\bibitem[\protect\citeauthoryear{Huang \bgroup et al\mbox.\egroup
  }{2017}]{Huang_2017_CVPR}
Huang, J.; Rathod, V.; Sun, C.; Zhu, M.; Korattikara, A.; Fathi, A.; Fischer,
  I.; Wojna, Z.; Song, Y.; Guadarrama, S.; and Murphy, K.
\newblock 2017.
\newblock Speed/accuracy trade-offs for modern convolutional object detectors.
\newblock In {\em 2017 IEEE Conference on Computer Vision and Pattern
  Recognition (CVPR)},  3296--3297.

\bibitem[\protect\citeauthoryear{Hutt}{1932}]{hutt1932kohs}
Hutt, M.
\newblock 1932.
\newblock The {K}ohs block-design tests: A revision for clinical practice.
\newblock {\em J. Applied Psychology} 16(3):298--307.

\bibitem[\protect\citeauthoryear{Jones and Torgesen}{1981}]{jones1981analysis}
Jones, R.~S., and Torgesen, J.~K.
\newblock 1981.
\newblock Analysis of behaviors involved in performance of the block design
  subtest of the {WISC-R}.
\newblock {\em Intelligence} 5(4):321--328.

\bibitem[\protect\citeauthoryear{Joy \bgroup et al\mbox.\egroup
  }{2001}]{joy2001quantifying}
Joy, S.; Fein, D.; Kaplan, E.; and Freedman, M.
\newblock 2001.
\newblock Quantifying qualitative features of block design performance among
  healthy older adults.
\newblock {\em Archives of Clinical Neuropsychology} 16(2):157--170.

\bibitem[\protect\citeauthoryear{Kanungo \bgroup et al\mbox.\egroup
  }{2000}]{Kanungo00anefficient}
Kanungo, T.; Mount, D.~M.; Netanyahu, N.~S.; Piatko, C.; Silverman, R.; and Wu,
  A.~Y.
\newblock 2000.
\newblock An efficient k-means clustering algorithm: Analysis and
  implementation.

\bibitem[\protect\citeauthoryear{Kendall}{1938}]{kendall1938measure}
Kendall, M.~G.
\newblock 1938.
\newblock A new measure of rank correlation.
\newblock {\em Biometrika} 30(1/2):81--93.

\bibitem[\protect\citeauthoryear{Kohs}{1920}]{kohs1920block}
Kohs, S.
\newblock 1920.
\newblock The block-design tests.
\newblock {\em J. Experimental Psychology} 3(5):357--376.

\bibitem[\protect\citeauthoryear{Kramer \bgroup et al\mbox.\egroup
  }{1991}]{kramer1991visual}
Kramer, J.; Blusewicz, M.; Kaplan, E.; and Preston, K.
\newblock 1991.
\newblock Visual hierarchical analysis of block design configural errors.
\newblock {\em J. Clin. and Exp. Neuropsychology} 13(4):455--465.

\bibitem[\protect\citeauthoryear{Kramer \bgroup et al\mbox.\egroup
  }{1999}]{kramer1999configural}
Kramer, J.~H.; Kaplan, E.; Share, L.; and Huckeba, W.
\newblock 1999.
\newblock Configural errors on {WISC-III} block design.
\newblock {\em J. International Neuropsychological Society} 5(6):518--524.

\bibitem[\protect\citeauthoryear{Lezak}{2004}]{lezak2004neuropsychological}
Lezak, M.~D.
\newblock 2004.
\newblock {\em Neuropsychological Assessment}.
\newblock Oxford University Press, USA.

\bibitem[\protect\citeauthoryear{Liu \bgroup et al\mbox.\egroup
  }{2015}]{Liu2017_ssd}
Liu, W.; Anguelov, D.; Erhan, D.; Szegedy, C.; Reed, S.~E.; Fu, C.; and Berg,
  A.~C.
\newblock 2015.
\newblock {SSD:} single shot multibox detector.
\newblock {\em CoRR} abs/1512.02325.

\bibitem[\protect\citeauthoryear{Milberg \bgroup et al\mbox.\egroup
  }{2009}]{milberg2009boston}
Milberg, W.~P.; Hebben, N.; Kaplan, E.; Grant, I.; and Adams, K.
\newblock 2009.
\newblock The boston process approach to neuropsychological assessment.
\newblock {\em Neuropsychological assessment of neuropsychiatric and
  neuromedical disorders}  42--65.

\bibitem[\protect\citeauthoryear{Miller \bgroup et al\mbox.\egroup
  }{2009}]{miller2009learning}
Miller, J.~C.; Ruthig, J.~C.; Bradley, A.~R.; Wise, R.~A.; Pedersen, H.~A.; and
  Ellison, J.~M.
\newblock 2009.
\newblock Learning effects in the block design task: A stimulus parameter-based
  approach.
\newblock {\em Psychological assessment} 21(4):570.

\bibitem[\protect\citeauthoryear{Qian \bgroup et al\mbox.\egroup
  }{2004}]{Qian:2004:SEC:967900.968151}
Qian, G.; Sural, S.; Gu, Y.; and Pramanik, S.
\newblock 2004.
\newblock Similarity between euclidean and cosine angle distance for nearest
  neighbor queries.
\newblock In {\em Proceedings of the 2004 ACM Symposium on Applied Computing},
  SAC '04,  1232--1237.
\newblock New York, NY, USA: ACM.

\bibitem[\protect\citeauthoryear{Rozencwajg and
  Corroyer}{2002}]{rozencwajg2002strategy}
Rozencwajg, P., and Corroyer, D.
\newblock 2002.
\newblock Strategy development in a block design task.
\newblock {\em Intelligence} 30(1):1--25.

\bibitem[\protect\citeauthoryear{Schatz, Ballantyne, and
  Trauner}{2000}]{schatz2000hierarchical}
Schatz, A.; Ballantyne, A.; and Trauner, D.
\newblock 2000.
\newblock A hierarchical analysis of block design errors in children with early
  focal brain damage.
\newblock {\em Dev. Neuropsychology} 17(1):75--83.

\bibitem[\protect\citeauthoryear{Stewart \bgroup et al\mbox.\egroup
  }{2009}]{stewart2009autistic}
Stewart, M.~E.; Watson, J.; Allcock, A.-J.; and Yaqoob, T.
\newblock 2009.
\newblock Autistic traits predict performance on the block design.
\newblock {\em Autism} 13(2):133--142.

\bibitem[\protect\citeauthoryear{Toraldo and Shallice}{2004}]{toraldo2004error}
Toraldo, A., and Shallice, T.
\newblock 2004.
\newblock Error analysis at the level of single moves in block design.
\newblock {\em Cognitive Neuropsychology} 21(6):645--659.

\bibitem[\protect\citeauthoryear{Troyer \bgroup et al\mbox.\egroup
  }{1994}]{troyer1994age}
Troyer, A.; Cullum, C.; Smernoff, E.; and Kozora, E.
\newblock 1994.
\newblock Age effects on block design: Qualitative performance features and
  extended-time effects.
\newblock {\em Neuropsychol.} 8:95--99.

\bibitem[\protect\citeauthoryear{Wechsler}{2003}]{wechsler2003wechsler}
Wechsler, D.
\newblock 2003.
\newblock {\em Wechsler intelligence scale for children ({WISC-IV})}.
\newblock The Psychological Corporation.

\bibitem[\protect\citeauthoryear{Wittenburg \bgroup et al\mbox.\egroup
  }{2006}]{wittenburg2006elan}
Wittenburg, P.; Brugman, H.; Russel, A.; Klassmann, A.; and Sloetjes, H.
\newblock 2006.
\newblock Elan: a professional framework for multimodality research.
\newblock In {\em 5th International Conference on Language Resources and
  Evaluation (LREC 2006)},  1556--1559.

\bibitem[\protect\citeauthoryear{Zipf-Williams \bgroup et al\mbox.\egroup
  }{2000}]{zipf2000qualitative}
Zipf-Williams, E.; Shear, P.; Strongin, D.; Winegarden, B.; and Morrell, M.
\newblock 2000.
\newblock Qualitative block design performance in epilepsy patients.
\newblock {\em Archives of Clinical Neuropsychology} 15:149--157.

\end{thebibliography}
\bibliographystyle{aaai}

\end{document}